# A multivalued knowledge-base model

Ágnes Achs
University of Pécs
Faculty of Engineering
Department of Computer Science
email: achs@pmmk.pte.hu

**Abstract.** The basic aim of our study is to give a possible model for handling uncertain information. This model is worked out in the framework of DATALOG. At first the concept of fuzzy Datalog will be summarized, then its extensions for intuitionistic- and interval-valued fuzzy logic is given and the concept of bipolar fuzzy Datalog is introduced. Based on these ideas the concept of multivalued knowledge-base will be defined as a quadruple of any background knowledge; a deduction mechanism; a connecting algorithm, and a function set of the program, which help us to determine the uncertainty levels of the results. At last a possible evaluation strategy is given.

## 1 Introduction

The dominant portion of human knowledge can not be modelled by pure inference systems, because this knowledge is often ambiguous, incomplete and vague. Several and often very different approaches have been used to study the inference systems. When knowledge is represented as a set of facts and rules, this uncertainty can be handled by means of fuzzy logic.

A few years ago in [1, 2] a possible combination of Datalog-like languages and fuzzy logic was presented. In these works the concept of fuzzy Datalog has been introduced by completing the Datalog-rules and facts by an uncertainty level and an implication operator. The level of a rule-head can be inferred from the level of the body and the level of the rule by the implication operator of the rule. Based upon our previous works, later on a fuzzy knowledge-base was developed, which is a possible background of an agent-model [3]. In the last years new steps were taken into the direction of multivalued knowledge-base: the fuzzy Datalog was extended to intuitionistic- and interval-valued fuzzy logic and the concept of bipolar fuzzy Datalog was introduced [4, 6]. In this year the concept of fuzzy knowledgebase was generalized into multivalued direction [7]

This paper wants to give an overview of the knowledge-bases based on fuzzy- or multivalued Datalog. In the first part the fuzzy, the intuitionistic, the interval-valued and the bipolar extensions of Datalog will be summarized. In the second part the concept of a possible multivalued knowledge-base will be discussed. This knowledge-base is a quadruple of a deduction mechanism; a background knowledge; an algorithm connecting the deduction and the knowledge; and a decoding set computing the uncertainty levels of the consequences. At last a possible evaluation of knowledge-bases will be shown.

## 2 Extensions of Datalog

Datalog is a logical programming language designed for use as a data-base query language.

A Datalog program consists of facts and rules. Using these rules new facts can be inferred from the program's facts. It is very important that the solution of a program be logically correct. This means that evaluating the program, the result be a model of the first order logic formulas, being its rules. On the other hand it is also important that this model would contain only those true facts which are the consequences of the program, that is the minimality of this model is expected, i.e. in this model it is impossible to make any true fact false and still have a model consistent with the database. An interpretation assigns truth or falsehood to every possible instance of the program's predicates. An interpretation is a model, if it makes the rules true, no matter what assignment of values from the domain is made for the variables in each rule. Although there are infinite many implications, it is proved that it is enough to consider only the Herbrand interpretation defined on the Herbrand universe and the Herbrand base.

The Herbrand universe of a program $\mathsf{P}$ (denoted by $\mathsf{H}_\mathsf{P}$) is the set of all possible ground terms constructed by using constants and function symbols occurring in $\mathsf{P}$. The Herbrand base of $\mathsf{P}$ ($\mathsf{B}_\mathsf{P}$) is the set of all possible ground atoms whose predicate symbols occur in $P$ and whose arguments are elements of $H_P$.

In general a term is a variable, a constant or a complex term of the form $f(t_1, \ldots, t_n)$, where f is a function symbol and $t_1, \ldots, t_n$ are terms. An atom is a formula of the form $p(t)$, where $p$ is a predicate symbol of a finite arity (say $n$) and $t$ is a sequence of terms of length $n$ (arguments). A literal is either an atom (positive literal) or its negation (negative literal). A term, atom or literal is ground if it is free of variables. As in fuzzy extension we did not deal with function symbols, so in our case the ground terms are the constants of the program.

In the case of Datalog programs there are several equivalent approaches to define the semantics of the program. In fuzzy extension we mainly rely on the fixed-point base aspect. The above concepts are detailed in classical works such as [15, 20, 21].

## 2.1 Fuzzy Datalog

In fuzzy Datalog (fDATALOG) we can complete the facts with an uncertainty level, the rules with an uncertainty level and an implication operator. We can infer for the level of a rule-head from the level of the rule-body and the level of the rule by the implication operator of the rule. As in classical cases, the logical correctness is extremely important as well, i.e., the solution would be a model of the program. This means that for each rule of the program, evaluating the fuzzy implication connecting to the rule, its truth-value has to be at least as large as the given uncertainty level. More precisely, the notion of fuzzy rule is the following:

An fDATALOG rule is a triplet $r; \beta; I$, where r is a formula of the form

$$A \leftarrow A_1, \ldots, A_n (n \geq 0),$$

$A$ is an atom (the head of the rule), $A_1, \ldots, A_n$ are literals (the body of the rule); I is an implication operator and $\beta \in (0, 1]$ (the level of the rule).

For getting a finite result, all the rules in the program must be safe. An fDATALOG rule is safe if all variables occurring in the head also occur in the body, and all variables occurring in a negative literal also occur in a positive one. An fDATALOG program is a finite set of safe fDATALOG rules.

There is a special type of rule, called fact. A fact has the form $A \leftarrow; \beta; I$. From now on, we refer to facts as $(A, \beta)$, because according to implication I, the level of $A$ can easily be computed and in the case of the implication operators detailed in this paper it is $\beta$.

For defining the meaning of a program, we need again the concepts of Herbrand universe and Herbrand base. Now a ground instance of a rule $r;\beta;I$ in $P$ is a rule obtained from $r$ by replacing every variable in $r$ with a constant of $H_P$. The set of all ground instances of $r;\beta;I$ is denoted by $ground(r);\beta;I$. The ground instance of $P$ is $ground(P) = \cup_{(r;\beta;I)\in P}(ground(r);\beta;I)$.

An interpretation of a program $P$ is a fuzzy set of the program's Herbrand base, $B_P$, i.e. it is: $\cup_{A\in B_P}(A,\alpha_A)$. An interpretation is a model of $P$ if for each $(ground(r);\beta;I) \in ground(P)$, $ground(r) = A \leftarrow A_1,\dots,A_n$

$$I(\alpha_{A_1\wedge\ldots\wedge A_n},\alpha_A) \geq \beta$$

A model $M$ is least if for any model $N$, $M \leq N$. A model $M$ is minimal if there is no model $N$, where $N \leq M$.

To be short we sometimes denote $\alpha_{A_1\wedge\ldots\wedge A_n}$, by $\alpha_{body}$ and $\alpha_A$ by $\alpha_{head}$.

In the extensions of Datalog several implication operators are used, but in all cases we are restricted to min-max conjunction and disjunction, and to the complement to 1 as negation. So: $\alpha_{A\wedge B} = min(\alpha_A,\alpha_B)$, $\alpha_{A\vee B} = max(\alpha_A,\alpha_B)$ and $\alpha_{\neg A} = 1-\alpha_A$.

The semantics of fDATALOG is defined as the fixed points of consequence transformations. Depending on these transformations, two semantics can be defined [1]. The deterministic semantics is the least fixed point of the deterministic transformation $DT_P$, the nondeterministic semantics is the least fixed point of the nondeterministic transformation $NT_P$. According to the deterministic transformation, the rules of a program are evaluated in parallel, while in the nondeterministic case the rules are considered independently and sequentially. These transformations are the following:

The $ground(P)$ is the set of all possible rules of $P$ the variables of which are replaced by ground terms of the Herbrand universe of $P$. $|A_i|$ denotes the kernel of the literal $A_i$, (i.e., it is the ground atom $A_i$, if $A_i$ is a positive literal, and $\neg A_i$, if $A_i$ is negative) and $\alpha_{body} = \min(\alpha_{A_1},\dots,\alpha_{A_n})$.

Let $B_P$ be the Herbrand base of the program $P$, and let $F(B_P)$ denote the set of all fuzzy sets over $B_P$. The consequence transformations $DT_P : F(B_P) \to F(B_P)$ and $NT_P : F(B_P) \to F(B_P)$ are defined as

$$DT_P(X) = (\cup_{R_A\in ground(P)}\{(A,\alpha_A)\}) \cup X \tag{1}$$

and

$$NT_P(X) = \{(A,\alpha_A)\} \cup X, \tag{2}$$

where $R_A : (A \leftarrow A_1,\dots,A_n;\beta;I) \in ground(P)$, $(|A_i|,\alpha_{A_i}) \in X$, $1 \leq i \leq n$; $\alpha_A = \max(0, \min\{\gamma \mid I(\alpha_{body},\gamma) \geq \beta\})$. $|A_i|$ denotes the kernel of the literal $A_i$, (i.e., it is the ground atom $A_i$, if $A_i$ is a positive literal, and $\neg A_i$, if $A_i$ is negative) and $\alpha_{body} = \min(\alpha_{A_1}, \dots, \alpha_{A_n})$.

In [1] it is proved that starting from the set of facts, both $DT_P$ and $NT_P$ have fixed points which are the least fixed points in the case of positive P. These fixed points are denoted by $lfp(DT_P)$ and $lfp(NT_P)$. It was also proved, that $lfp(DT_P)$ and $lfp(NT_P)$ are models of P, so we could define $lfp(DT_P)$ as the deterministic semantics, and $lfp(NT_P)$ as the nondeterministic semantics of fDATALOG programs. For a function- and negation-free fDATALOG, the two semantics are the same, but they are different if the program has any negation. In this case the set $lfp(DT_P)$ is not always a minimal model, but the nondeterministic semantics – $lfp(NT_P)$ – is minimal under certain conditions. These conditions are referred to as stratification. Stratification gives an evaluating sequence in which the negative literals are evaluated first [2].

To compute the level of rule-heads, we need the concept of the uncertainty-level function, which is:

$$f(I, \alpha, \beta) = \min(\{\gamma \mid I(\alpha, \gamma) \geq \beta\}).$$

According to this function the level of a rule-head is: $\alpha_{head}= f(I, \alpha_{body}, \beta)$.

In the former papers [1, 2] several implications were detailed (the operators treated in [17]), and the conditions of the existence of an uncertainty-level function was examined for all these operators. For intuitionistic cases three of them is extended in this paper. They are the following:

| | | |
|---|---|---|
| Gödel | $I_G(\alpha, \gamma) = \begin{cases} 1 & \alpha \leq \gamma \\ \gamma & \text{otherwise} \end{cases}$ | $f(I_G, \alpha, \beta) = \min(\alpha, \beta)$ |
| Lukasiewicz | $I_L(\alpha, \gamma) = \begin{cases} 1 & \alpha \leq \gamma \\ 1 - \alpha + \gamma & \text{otherwise} \end{cases}$ | $f(I_L, \alpha, \beta) = \max(0, \alpha + \beta - 1)$ |
| Kleene-Dienes | $I_K(\alpha, \gamma) = \max(1 - \alpha, \gamma)$ | $f(I_K, \alpha, \beta) = \begin{cases} 0 & \alpha + \beta \leq 1 \\ \beta & \alpha + \beta > 1 \end{cases}$ |

**Example 1** *Let us consider the next program:*

$$\begin{aligned}
&(p(a), 0.8).\\
&(r(b), 0.6).\\
&s(x) \quad \leftarrow \ q(x,y); 0.7; I_L.\\
&q(x,y) \ \leftarrow \ p(x), r(y); 0.7; I_G.\\
&q(x,y) \ \leftarrow \ \neg q(y,x); 0.9; I_K.
\end{aligned}$$

As the program has a negation, so according to the stratification the right order of rule-evaluation is 2.,3,.1. Then

$$\begin{aligned}
&lfp(NT_P) =\\
&\{(p(a), 0.8); (r(b), 0.6); (q(a,b), 0.6);\\
&(q(b,a), 0.9); (s(a), 0.3); (s(b), 0.6)\}.
\end{aligned}$$

## 2.2 Multivalued extensions of fuzzy Datalog

In fuzzy set theory the membership of an element in a fuzzy set is a single value between zero and one, and the degree of non-membership is automatically just the complement to 1 of the membership degree. However a human being who expresses the degree of membership of a given element in a fuzzy set, very often does not express a corresponding degree of non-membership as its complement. That is, there may be some hesitation degree. This illuminates a well-known psychological fact that linguistic negation does not always correspond to logical negation. Because of this observation, as a generalization of fuzzy sets, the concept of intuitionistic fuzzy sets was introduced by Atanassov in 1983 [9, 11]. In the next paragraphs some possible multivalued extensions will be discussed.

### 2.2.1 Intuitionistic and interval-valued extensions of fuzzy Datalog

While in fuzzy logic the uncertainty is represented by a single value ($\mu$), in intuitionistic-(IFS) and interval-valued (IVS) fuzzy logic it is represented by two values, $\vec{\mu} = (\mu_1, \mu_2)$. In the intuitionistic case the two elements must satisfy the condition $\mu_1 + \mu_2 \leq 1$, while in the interval-valued case the condition is $\mu_1 \leq \mu_2$. In IFS $\mu_1$ is the degree of membership and $\mu_2$ is the degree of non-membership, while in IVS the membership degree is between $\mu_1$ and $\mu_2$. It is obvious that the relations $\mu_1' = \mu_1$, $\mu_2' = 1 - \mu_2$ create a mutual connection between the two systems. (The equivalence of IVS and IFS was stated first in [10].) In both cases an ordering relation can be defined, and according to this ordering a lattice is taking shape:

$L_F$ and $L_V$ are lattices of IFS and IVS respectively, where:

$$L_F = \{(x_1, x_2) \in [0,1]^2 \mid x_1 + x_2 \leq 1\},$$
$$(x_1, x_2) \leq_F (y_1, y_2) \Leftrightarrow x_1 \leq y_1, x_2 \geq y_2$$

$$L_V = \{(x_1, x_2) \in [0,1]^2 \mid x_1 \leq x_2\},$$
$$(x_1, x_2) \leq_V (y_1, y_2) \Leftrightarrow x_1 \leq y_1, x_2 \leq y_2.$$

It can be proved that both $L_F$ and $L_V$ are complete lattices [16]. In both cases the extended fDATALOG is defined on these lattices and the necessary concepts are generalizations of the ones presented above.

**Definition 2** *The i-extended fDATALOG program (ifDATALOG) is a finite set of safe ifDATALOG rules* $(r; \vec{\beta}; \vec{I}_{FV})$*;*

- *the i-extended consequence transformations* $iDT_P$ *and* $iNT_P$ *are formally the same as* $DT_P$ *and* $NT_P$ *in (1), (2) except:* $\vec{\alpha}_A = \max(\vec{0}_{FV}, \min\{\vec{\gamma} \mid \vec{I}_{FV}(\vec{\alpha}_{body}, \vec{\gamma}) \geq_{FV} \vec{\beta}\})$ *and*
- *the* $i$*-extended uncertainty-level function is* $f(\vec{I}_{FV}, \vec{\alpha}, \vec{\beta}) = \min(\{\vec{\gamma} \mid \vec{I}_{FV}(\vec{\alpha}, \vec{\gamma}) \geq_{FV} \vec{\beta}\})$,

*where* $\vec{\alpha}$, $\vec{\beta}$, $\vec{\gamma}$ *are elements of* $L_F$, $L_V$ *respectively,* $\vec{I}_{FV} = \vec{I}_F$ *or* $\vec{I}_V$ *is an implication of* $L_F$ *or* $L_V$*,* $\vec{0}_{FV}$ *is* $\vec{0}_F = (0,1)$ *or* $\vec{0}_V = (0,0)$ *and* $\geq_{FV}$ *is* $\geq_F$ *or* $\geq_V$*.*

As $iDT_P$ and $iNT_P$ are inflationary transformations over the complete lattices $L_F$ or $L_V$, thus according to [15] they have an inflationary fixed point denoted by $lfp(iDT_P)$ and $lfp(iNT_P)$. If $P$ is positive (without negation), then $iDT_P = iNT_P$ is a monotone transformation, so $lfp(iDT_P) = lfp(iNT_P)$ is the least fixed point.

The fixed point is an interpretation of $P$, which is a model, if for each

$$(A \leftarrow A_1, \ldots, A_n; \vec{\beta}; \vec{I}_{FV}) \in ground(P), \quad \vec{I}_{FV}(\vec{\alpha}_{body}, \vec{\alpha}_A) \geq_{FV} \vec{\beta}.$$

It can easily be proved that these fixed points are models of the program.

**Proposition 3** $lfp(iDT_P)$ *and* $lfp(iNT_P)$ *are models of* $P$*.*

**Proof.** For $T = iDT_P$ or $T = iNT_P$ there are two kinds of rules in $ground(P)$: a/ $A \leftarrow A_1, \ldots, A_n; \vec{\beta}; \vec{I}_{FV}$; $(A, \vec{\alpha}_A) \in lfp(T)$; $(|A_i|, \vec{\alpha}_{A_i}) \in lfp(T)$, $1 \leq i \leq n$.

b/ $A \leftarrow A_1, \ldots, A_n; \vec{\beta}; \vec{I}_{FV}; \exists i : (|A_i|, \vec{\alpha}_{A_i}) \notin \mathrm{lfp}(T)$.
In the first case according to the construction of $\vec{\alpha}_A$, $\vec{I}_{FV}(\vec{\alpha}_{body}, \vec{\alpha}_A) \geq_{FV} \vec{\beta}$ holds, in the second case $A_i$ is not among the facts, so $\vec{\alpha}_{A_i} = \vec{0}_{FV}$, therefore $\vec{\alpha}_{body} = \vec{0}_{FV}$ and $\vec{I}_{FV}(\vec{\alpha}_{body}, \vec{\alpha}_A) = \vec{1}_{FV} \geq_{FV} \vec{\beta}$. That is $\mathrm{lfp}(T)$ is a model. □

According to the above statements the next theorem is true:

**Theorem 4** *Both* $\mathrm{iDT_P}$ *and* $\mathrm{iNT_P}$ *have a fixed point, denoted by* $\mathrm{lfp}(\mathrm{iDT_P})$ *and* $\mathrm{lfp}(\mathrm{iNT_P})$*. If* $P$ *is positive, then* $\mathrm{lfp}(\mathrm{iDT_P}) = \mathrm{lfp}(\mathrm{iNT_P})$ *and this is the least fixed point.* $\mathrm{lfp}(\mathrm{iDT_P})$ *and* $\mathrm{lfp}(\mathrm{iNT_P})$ *are models of* $P$*; for negation-free ifDATALOG this is the least model of the program.*

As the intuitionistic or interval-valued extension of Datalog has no influence on the stratification, the propositions detailed in the case of stratified fDATALOG programs are true in the case of $i$-extended fuzzy Datalog programs as well:

**Proposition 5** *For stratified ifDATALOG program* $P$*, there is an evaluation sequence, in which* $\mathrm{lfp}(\mathrm{iNT_P})$ *is a unique minimal model of* $P$*.*

After defining the syntax and semantics of $i$-extended fuzzy Datalog, it is necessary to examine the properties of possible implication operators and the $i$-extended uncertainty-level functions. A number of intuitionistic implications are established in [16, 12, 13] and other papers, four of which are the extensions of the above three fuzzy implication operators thus chosen for now. For these operators the suitable interval-valued operators will be decided, and for both kinds of them we will deduce the uncertainty-level functions. Now the computations will not be shown, only the starting points and results are presented.

The coordinates of intuitionistic and interval-valued implication operators can be determined by each other. The uncertainty-level functions can be computed according to the applied implication. The connection between $I_F$ and $I_V$ and the extended versions of uncertainty-level functions are given below:

$$\vec{I}_V(\vec{\alpha}, \vec{\gamma}) = (I_{V1}, I_{V2}) \text{ where}$$

$$\begin{aligned} I_{V1} &= I_{F1}(\vec{\alpha}', \vec{\gamma}'), & \vec{\alpha}' &= (\alpha_1, 1-\alpha_2), \\ I_{V2} &= 1 - I_{F2}(\vec{\alpha}', \vec{\gamma}')); & \vec{\gamma}' &= (\gamma_1, 1-\gamma_2). \end{aligned}$$

$$f(\vec{I}_F, \vec{\alpha}, \vec{\beta}) = (\min(\{\gamma_1 \mid I_{F1}(\vec{\alpha}, \vec{\gamma}) \geq \beta_1\}), \max(\{\gamma_2 \mid I_{F2}(\vec{\alpha}, \vec{\gamma}) \leq \beta_2\}));$$

$$f(\vec{I}_V, \vec{\alpha}, \vec{\beta}) = (\min(\{\gamma_1 \mid I_{V1}(\vec{\alpha}, \vec{\gamma}) \geq \beta_1\}), \min(\{\gamma_2 \mid I_{V2}(\vec{\alpha}, \vec{\gamma}) \geq \beta_2\})).$$

The studied operators and the related uncertainty-level functions are the following:

**Extension of the Kleene-Dienes implication**

One possible extension of the Kleene-Dienes implication for IFS is:

$$\vec{I}_{FK}(\vec{\alpha}, \vec{\gamma}) = (\max(\alpha_2, \gamma_1), \min(\alpha_1, \gamma_2)).$$

The appropriate computed elements are:

$$\vec{I}_{VK}(\vec{\alpha}, \vec{\gamma}) = (\max(1 - \alpha_2, \gamma_1), \max(1 - \alpha_1, \gamma_2));$$

$$f_1(\vec{I}_{FK}, \vec{\alpha}, \vec{\beta}) = \begin{cases} 0 & \alpha_2 \geq \beta_1 \\ \beta_1 & \text{otherwise} \end{cases}, \quad f_1(\vec{I}_{VK}, \vec{\alpha}, \vec{\beta}) = \begin{cases} 0 & 1 - \alpha_2 \geq \beta_1 \\ \beta_1 & \text{otherwise} \end{cases},$$

$$f_2(\vec{I}_{FK}, \vec{\alpha}, \vec{\beta}) = \begin{cases} 1 & \alpha_1 \leq \beta_2 \\ \beta_2 & \text{otherwise} \end{cases}, \quad f_2(\vec{I}_{VK}, \vec{\alpha}, \vec{\beta}) = \begin{cases} 0 & (1 - \alpha_1 \leq \beta_2 \\ \beta_2 & \text{otherwise} \end{cases}.$$

**Extension of the Lukasiewicz implication**

One possible extension of the Lukasiewicz implication for IFS is:

$$\vec{I}_{FL}(\vec{\alpha}, \vec{\gamma}) = (\min(1, \alpha_2 + \gamma_1), \max(0, \alpha_1 + \gamma_2 - 1)).$$

The appropriate computed elements are:

$$\vec{I}_{VL}(\vec{\alpha}, \vec{\gamma}) = (\min(1, 1 - \alpha_2 + \gamma_1), \min(1, 1 - \alpha_1 + \gamma_2));$$

$$\begin{aligned} f_1(\vec{I}_{FK}, \vec{\alpha}, \vec{\beta}) &= \min(1 - \alpha_2, \max(0, \beta_1 - \alpha_2)), \\ f_2(\vec{I}_{FK}, \vec{\alpha}, \vec{\beta}) &= \max(1 - \alpha_1, \min(1, 1 - \alpha_1 + \beta_2)); \end{aligned}$$

$$\begin{aligned} f_1(\vec{I}_{VK}, \vec{\alpha}, \vec{\beta}) &= \max(0, \alpha_2 + \beta_1 - 1), \\ f_2(\vec{I}_{VK}, \vec{\alpha}, \vec{\beta}) &= \max(0, \alpha_1 + \beta_2 - 1). \end{aligned}$$

**Extension of the Gödel implication**

There are several alternative extensions of the Gödel implication, now we present two of them:

$$\vec{I}_{FG1}(\vec{\alpha},\vec{\gamma})=\begin{cases}(1,0) & \alpha_1 \le \gamma_1,\\ (\gamma_1,0) & \alpha_1 > \gamma_1, \alpha_2 \ge \gamma_2,\\ (\gamma_1,\gamma_2) & \alpha_1 > \gamma_1, \alpha_2 < \gamma_2,\end{cases}$$

$$\vec{I}_{FG2}(\vec{\alpha},\vec{\gamma})=\begin{cases}(1,0) & \alpha_1 \le \gamma_1, \alpha_2 \ge \gamma_2,\\ (\gamma_1,\gamma_2) & \text{otherwise.}\end{cases}$$

The appropriate computed elements are:

$$\vec{I}_{VG1}(\vec{\alpha},\vec{\gamma})=\begin{cases}(1,1) & \alpha_1 \le \gamma_1,\\ (\gamma_1,1) & \alpha_1 > \gamma_1, \alpha_2 \ge \gamma_2,\\ (\gamma_1,\gamma_2) & \alpha_1 > \gamma_1, \alpha_2 < \gamma_2,\end{cases}$$

$$\vec{I}_{VG2}(\vec{\alpha},\vec{\gamma})=\begin{cases}(1,1) & \alpha_1 \le \gamma_1, \alpha_2 \le \gamma_2,\\ (\gamma_1,\gamma_2) & \text{otherwise,}\end{cases}$$

$$f_1(\vec{I}_{FG1},\vec{\alpha},\vec{\beta}) = \min(\alpha_1,\beta_1), \qquad f_1(\vec{I}_{FG2},\vec{\alpha},\vec{\beta}) = \min(\alpha_1,\beta_1),$$

$$f_2(\vec{I}_{FG1},\vec{\alpha},\vec{\beta}) = \begin{cases}1 & \alpha_1 \le \beta_1,\\ \max(\alpha_2,\beta_2) & \text{otherwise;}\end{cases} \qquad f_2(\vec{I}_{FG2},\vec{\alpha},\vec{\beta}) = \max(\alpha_2,\beta_2),$$

$$f_1(\vec{I}_{VG1},\vec{\alpha},\vec{\beta}) = \min(\alpha_1,\beta_1), \qquad f_1(\vec{I}_{VG2},\vec{\alpha},\vec{\beta}) = \min(\alpha_1,\beta_1),$$

$$f_2(\vec{I}_{VG1},\vec{\alpha},\vec{\beta}) = \begin{cases}0 & \alpha_1 \le \beta_1,\\ \min(\alpha_2,\beta_2) & \text{otherwise;}\end{cases} \qquad f_2(\vec{I}_{VG2},\vec{\alpha},\vec{\beta}) = \min(\alpha_2,\beta_2).$$

An extremely important question is whether the consequences of the program remain within the scope of intuitionistic or interval-valued fuzzy logic. That is if the levels of the body and the rule satisfy the conditions referring to intuitionistic or interval-valued concepts, does the resulting level of the head also satisfy these conditions?

Unfortunately for implications other than G2, the resulting degrees do not fulfill these conditions in all cases. For example in the case of the Kleene-Dienes and the Lukasiewicz intuitionistic operators the condition of intuitionism satisfies for the levels of the rule-head, only if the sum of the levels of the rule-body is at least as large as the sum of the levels of the rule:

$$f_1(\vec{I}_F, \vec{\alpha}, \vec{\beta}) + f_2(\vec{I}_F, \vec{\alpha}, \vec{\beta}) \leq 1 \text{ if } \alpha_1 + \alpha_2 \geq \beta_1 + \beta_2.$$

That is the solution is inside the scope of IFS, if the level of the rule-body is less "intuitionistic" than the level of the rule.

In the case of the first Gödel operator the condition is simpler:

$$f_1(\vec{I}_F, \vec{\alpha}, \vec{\beta}) + f_2(\vec{I}_F, \vec{\alpha}, \vec{\beta}) \leq 1 \text{ if } \alpha_1 > \beta_1,$$

i.e. the solution is inside the scope of IFS, only if the level of the rule-body is more certain than the level of the rule.

Maybe in practical cases these conditions are satisfied, but the examination of this question may be the subject of further research. For now the next proposition can easily be proved:

**Proposition 6** *For* $\vec{\alpha} = (\alpha_1, \alpha_2)$, $\vec{\beta} = (\beta_1, \beta_2)$
*if* $\alpha_1 + \alpha_2 \leq 1$, $\beta_1 + \beta_2 \leq 1$ *then* $f_1(\vec{I}_{FG2}, \vec{\alpha}, \vec{\beta}) + f_2(\vec{I}_{FG2}, \vec{\alpha}, \vec{\beta}) \leq 1$;
*if* $\alpha_1 \leq \alpha_2$, $\beta_1 \leq \beta_2$ *then* $f_1(\vec{I}_{VG2}, \vec{\alpha}, \vec{\beta}) \leq f_2(\vec{I}_{VG2}, \vec{\alpha}, \vec{\beta})$.

A further important question is whether the fixed-point algorithm terminates or not, that is whether or not the consequence transformations reach the fixed point in finite steps. As $P$ is finite, the fixed point contains only finite many elements. The only problem may occur with the uncertainty levels of recursive rules. It can be seen that the recursion terminates if $\vec{f}(\vec{I}_{FV}, \vec{\alpha}, \vec{\beta}) \leq_{FV} \vec{\alpha}$ for each $\vec{\alpha} \in L_{FV}$. As

$$\vec{f}(\vec{I}_{FG2}, \vec{\alpha}, \vec{\beta}) = (\min(\alpha_1, \beta_1), \max(\alpha_2, \beta_2)) \leq_F \vec{\alpha}$$

and

$$\vec{f}(\vec{I}_{VG2}, \vec{\alpha}, \vec{\beta}) = (\min(\alpha_1, \beta_1), \min(\alpha_2, \beta_2)) \leq_V \vec{\alpha}$$

G2 satisfies this condition, so:

**Proposition 7** *In the case of G2 operator the fixed-point algorithm terminates.*

### 2.2.2 Bipolar extension of fuzzy Datalog

The above mentioned problem of extended implications other than G2 and the results of certain psychological researches have led to the idea of bipolar fuzzy Datalog. The intuitive meaning of intuitionistic degrees is based on psychological observations, namely on the idea that concepts are more naturally approached through separately envisaging positive and negative instances [14, 18, 19]. Taking a further step, there are differences not only in the instances but also in the way of thinking as well. There is a difference between positive and negative thinking, between deducing positive or negative uncertainty. The idea of bipolar Datalog is based on the previous observation: we use two kinds of ordinary fuzzy implications for positive and negative inference, namely we define a pair of consequence transformations instead of a single one. Since in the original transformations lower bounds are used with degrees of uncertainty, therefore starting from IFS facts, the resulting degrees will be lower bounds of membership and non-membership respectively, instead of the upper bound for non-membership. However, if each non-membership value $\mu$ is transformed into membership value $\mu' = 1 - \mu$ , then both members of head-level can be inferred similarly. Therefore, two kinds of bipolar evaluations have been defined.

**Definition 8** *The bipolar fDATALOG program (bfDATALOG) is a finite set of safe bfDATALOG rules* $(r; (\beta_1, \beta_2); (I_1, I_2))$;

- *in variant "a" the elements of bipolar consequence transformations:* $b\vec{DT}_P = (DT_{P1}, DT_{P2})$ *and* $b\vec{NT}_P = (NT_{P1}, NT_{P2})$ *are the same as* $DT_P$ *and* $NT_P$ *in (1), (2),*
- *in variant "b" in* $DT_{P2}$ *and* $NT_{P2}$ *the level of rule's head is:* $\alpha'_{A2} = \max(0, \min\{\gamma'_2 | I_2(\alpha'_{body2}, \gamma'_2) \geq \beta'_2\})$; *where* $\alpha'_{body2} = \min(\alpha'_{A_1 2}, \ldots, \alpha'_{A_N 2})$

According to the two variants the uncertainty-level functions are:

$$\begin{aligned}
\vec{f}_a &= (f_{a1}, f_{a2}); \quad \vec{f}_b = (f_{b1}, f_{b2}); \\
f_{a1} &= f_{b1} = \min\{\gamma_1 \mid I_1(\alpha_1, \gamma_1) \geq \beta_1\}; \\
f_{a2} &= \min\{\gamma_2 \mid I_2(\alpha_2, \gamma_2) \geq \beta_2\}; \\
f_{b2} &= 1 - \min\{1 - \gamma_2 \mid I_2(1 - \alpha_2, 1 - \gamma_2) \geq 1 - \beta_2\}.
\end{aligned}$$

It is evident, that applying the transformation $\mu'_1 = \mu_1$, $\mu'_2 = 1 - \mu_2$, for each IFS levels of the program, variant "b" can be applied to IVS degrees as well. Contrary to the results of ifDATALOG, the resulting degrees of most variants of bipolar fuzzy Datalog satisfy the conditions referring to IFS and IVS respectively. A simple computation can prove the next proposition:

**Proposition 9** *For* $\vec{\alpha} = (\alpha_1, \alpha_2)$, $\vec{\beta} = (\beta_1, \beta_2)$ *and for* $(I_1, I_2) = (I_G, I_G)$; $(I_1, I_2) = (I_L, I_L)$; $(I_1, I_2) = (I_L, I_G)$; $(I_1, I_2) = (I_K, I_K)$; $(I_1, I_2) = (I_L, I_K)$

*if* $\alpha_1 + \alpha_2 \le 1,\ \beta_1 + \beta_2 \le 1$ *then* $f_{a1}(I_1, \vec{\alpha}, \vec{\beta}) + f_{a2}(I_2, \vec{\alpha}, \vec{\beta}) \le 1$
*and* $f_{b1}(I_1, \vec{\alpha}, \vec{\beta}) + f_{b2}(I_2, \vec{\alpha}, \vec{\beta}) \le 1$;

*further on*

$$f_{a1}(I_G, \vec{\alpha}, \vec{\beta}) + f_{a2}(I_L, \vec{\alpha}, \vec{\beta}) \le 1; \qquad f_{a1}(I_G, \vec{\alpha}, \vec{\beta}) + f_{a2}(I_K, \vec{\alpha}, \vec{\beta}) \le 1;$$
$$f_{a1}(I_K, \vec{\alpha}, \vec{\beta}) + f_{a2}(I_G, \vec{\alpha}, \vec{\beta}) \le 1; \qquad f_{a1}(I_K, \vec{\alpha}, \vec{\beta}) + f_{a2}(I_L, \vec{\alpha}, \vec{\beta}) \le 1.$$

Although there are more results for variant "a", it seems that the model realised by variant "b" is more useful.

Because of the construction of bipolar consequence transformations the following proposition is evident:

**Proposition 10** *Both variations of bipolar consequence transformations have a least fixed point, which are models of* $\mathsf{P}$ *in the following sense:*
*a/ for each*

$$(A \leftarrow A_1, \ldots, A_n; \vec{\beta}; \vec{I}) \in \mathsf{ground}(P)\ I(\alpha_{body1}, \alpha_{A1}) \ge \beta_1; I(\alpha_{body2}, \alpha_{A2}) \ge \beta_2,$$

*b/ for each*

$$(A \leftarrow A_1, \ldots, A_n; \vec{\beta}; \vec{I}) \in \mathsf{ground}(P)\ I(\alpha_{body1}, \alpha_{A1}) \ge \beta_1; I(\alpha'_{body2}, \alpha'_{A2}) \ge \beta'_2.$$

**Proof.** The termination of the consequence transformations based on these three implication operators was proved in the case of fDATALOG, and since this property does not change in bipolar case, the bipolar consequence transformations terminate as well. □

As the bipolar extension of Datalog has no influence on the stratification, therefore the propositions detailed in the case of stratified fDATALOG programs are true in the case of bipolar fuzzy Datalog programs as well:

**Proposition 11** *For stratified bfDATALOG program* $\mathsf{P}$, *there is an evaluation sequence, in which* $\mathsf{lfp}(b\vec{N}T_P)$ *is a unique minimal model of* $\mathsf{P}$.

**Example 12** *Let us consider the next program:*

$$(p(a,b),(0.6,0.2)). \qquad (p(a,c),(0.7,0.3)).$$

$$(p(b,d),(0.5,0.3)). \qquad (p(d,e),(0.8,0.1)).$$

$$q(x,y) \leftarrow p(x,y); \vec{I_1}; (0.75,0.2).$$
$$q(x,y) \leftarrow p(x,z), q(z,y); \vec{I_2}; (0.7,0.2).$$

*According to it uncertainty levels this program can be evaluated in intuitionistic or bipolar manner. At first let us see the intuitionistic evaluation.*

*Let* $\vec{I}_1 = \vec{I}_2 = \vec{I}_{FG2}(\vec{\alpha},\vec{\gamma}) = \begin{cases} (1,0) & \alpha_1 \le \gamma_1, \alpha_2 \ge \gamma_2 \\ (\gamma_1,\gamma_2) & \text{otherwise} \end{cases}$.
*Then* $f_1(\vec{I}_{FG2},\vec{\alpha},\vec{\beta}) = \min(\alpha_1,\beta_1), f_2(\vec{I}_{FG2},\vec{\alpha},\vec{\beta}) = \max(\alpha_2,\beta_2)$.

*Without regarding all of the details only three computations will be shown: from the first rule the uncertainty of* $q(a,b)$ *and* $q(b,d)$ *and from the second one the uncertainty of* $q(a,d)$.

*The uncertainty of* $q(a,b)$ *is:* $(\min(0.6,0.75),\max(0.2,0.2)) = (0.6,0.2)$. *Similarly the uncertainty of* $q(b,d)$ *is* $(0.5,0.3)$.

*In the case of* $q(a,d)$ *its uncertainty can be computed from the appropriate* $(\text{ground}(r);\beta;I)$*, where r is the second rule, that is ground(r) is* $q(a,d) \leftarrow p(a,b), q(b,d)$. *The uncertainty of the rule-body is* $\min_F((0.6,0.2),(0.5,0.3)) = (\min(0.6,0.5),\max(0.2,0.3)) = (0.5,0.3)$. *According to the uncertainty function the level of* $q(b,d)$ *is:* $(\min(0.5,0.7),\max(0.3,0.2)) = (0.5,0.3)$.

*Computing the other atoms, the fixed point is:*

$$\{(p(a,b),(0.6,0.2)),(p(a,c),(0.7,0.3)),(p(b,d),(0.5,0.3)),(p(d,e),(0.8,0.1)),$$
$$(q(a,b),(0.6,0.2)),(q(a,c),(0.7,0.3)),(q(b,d),(0.5,0.3)),(q(d,e),(0.75,0.2),$$
$$(q(a,d),(0.5,0.3)),(q(b,e),(0.5,0.3)),(q(a,e),(0.5,0.3))\}.$$

*Now let the program be evaluated in bipolar manner according to variant "b" and let* $\vec{I_1} = (I_L, I_K), \vec{I_2} = (I_G, I_G)$*, i.e.*

$$f(I_G,\alpha,\beta) = \min(\alpha,\beta),$$
$$f(I_L,\alpha,\beta) = \max(0,\alpha+\beta-1),$$
$$f(I_K,\alpha,\beta) = \begin{cases} 0 & \alpha+\beta \le 1, \\ \beta & \alpha+\beta > 1. \end{cases}$$

*Then the first coordinates of the computed uncertainties are:*

$$(q(a,b),(\max(0,0.6+0.75-1)=0.35,_)),(q(a,c),(0.45,_)),$$
$$(q(b,d),(0.25,_)),(q(d,e),(0.55,_)),$$
$$(q(a,d): \textit{as } \min(0.6,0.25)+0.7\le 1 \textit{ so } (q(a,d),(0,_)),$$
$$(q(b,e),(0.7,_)),(q(a,e),(0.7,_)).$$

*The second coordinates are computed after the appropriate transformation* $\alpha_2' = 1-\alpha_2$. *So*

$$(q(a,b),(_,1-\min(1-0.2,1-0.2)=0.2)),(q(a,c),(_,0.3)),(q(b,d),(_,0.3)),$$
$$(q(d,e),(_,0.2),(q(a,d),(_,0.3)),(q(b,e),(_,0.3)),(q(a,e),(_,0.3)).$$

*So the fixed point is:*

$$\{(p(a,b),(0.6,0.2)),(p(a,c),(0.7,0.3)),(p(b,d),(0.5,0.3)),$$
$$(p(d,e),(0.8,0.1)),(q(a,b),(0.35,0.2)),(q(a,c),(0.45,0.3)),$$
$$(q(b,d),(0.25,0.3)),(q(d,e),(0.55,0.2),(q(a,d),(0,0.3)),$$
$$(q(b,e),(0.7,0.3)),(q(a,e),(0.7,0.3))\}.$$

## 2.3 Evaluation of programs

The above examples show that the fixed point-query – that is the bottom-up evaluation – may involve many superfluous calculations, because sometimes we want to give an answer to a concrete question, and we are not interested in the whole sequence. If a goal is specified together with an fDATALOG (or ifDATALOG, bfDATALOG) program, it is enough to consider only the rules and facts necessary to reaching the goal.

Generally by the top down evaluation the goal is evaluated through subqueries. This means that all the possible rules whose head can be unified with the given goal are selected and the atoms of the body are considered as new sub-goals. This procedure continues until the facts are obtained. The evaluation of a fuzzy Datalog or multivalued Datalog program does not terminate by obtaining the facts, because we need to determine the uncertainty level of the goal. The evaluation has a second part; it calculates this level in a bottom-up manner: starting from the leaves of the evaluating graph, going backward to the root, and applying the uncertainty-level functions along the suitable path of this graph, finally we get the uncertainty level of the root.

For a fuzzy Datalog program a goal is a pair $(Q;\alpha_Q)$, where Q is an atom, $\alpha_Q$ is the uncertainty level of the atom. Q may contain variables, and its level may be known or unknown value. An fDATALOG program enlarged with a goal is a fuzzy query.

For a multivalued Datalog program the goal is very similar, except instead of one level, the goal-atom's belonging level is a pair of levels. An ifDATALOG or a bfDATALOG program enlarged with a goal is an intuitionistic-, interval-valued- or bipolar-query. As the evaluating algorithm applies the uncertainty level function independently of its meaning, therefore this algorithm is suitable for all discussed type of fuzzy Datalog or extended Datalog.

Now the evaluating algorithm of the queries will not be detailed because the aim of this paper is to build a knowledge-base system and the evaluation of a knowledgebase differs from the evaluation of a program. The evaluation algorithm of multivalued Datalog programs is discussed in [8].

# 3 Multivalued knowledge-base

As fuzzy Datalog is a special kind of its each multivalued extension, so further on both fDATALOG and any of above extensions will be called multivalued Datalog (mDATALOG).

## 3.1 Background knowledge

The facts and rules of an mDATALOG program can be regarded as any kind of knowledge, but sometimes we need some other information in order to get an answer for a query. In this section we give a possible model of background knowledge. Some kind of synomyms will be defined between the potential predicates and between the potential constans of the given problem, so it can be examined in a larger context. More precisely a proximity relation will be defined between predicates and between constants and these structures of proximity will serve as a background knowledge.

**Definition 13** *A multivalued proximity on a domain* $D$ *is an IFS or IVS valued relation* $\vec{R}_{FV_D} : D \times D \to [\vec{0}_{FV}, \vec{1}_{FV}]$ *which satisfies the following properties:*

$$\vec{R}_{F_D}(x,y) = \vec{\mu}_F(x,y) = (\mu_1, \mu_2), \quad \mu_1 + \mu_2 \le 1$$
$$\vec{R}_{V_D}(x,y) = \vec{\mu}_V(x,y) = (\mu_1, \mu_2), \quad 0 \le \mu_1 \le \mu_2 \le 1$$
$$\vec{R}_{FV_D}(x,x) = \vec{1}_{FV} \quad \forall x \in D \quad (reflexivity)$$
$$\vec{R}_{FV_D}(x,y) = \vec{R}_{FV_D}(y,x) \quad \forall x, y \in D \quad (symmetry).$$

A proximity is similarity if it is transitive, that is

$$\vec{R}_{FV_D}(x,z) \geq \min(\vec{R}_{FV_D}(x,y), \vec{R}_{FV_D}(y,z)) \ \forall x,y,z \in D.$$

In the case of similarity equivalence classifications can be defined over $D$ allowing to develop simpler or more effective algorithms, but now we deal with the more general proximity.

In our model the background knowledge is a set of proximity sets.

**Definition 14** *Let* $d \in D$ *any element of domain* $D$*. The proximity set of* $d$ *is an IFS or IVS subset over* $D$*:*

$$R_{FV_d} = \{(d_1, \vec{\lambda}_{FV_1}), (d_2, \vec{\lambda}_{FV_2}), \ldots, (d_n, \vec{\lambda}_{FV_n})\},$$

*where* $d_i \in D$ *and* $\vec{R}_{FV_D}(d, d_i) = \vec{\lambda}_{FV_i}$ *for* $i = 1, \ldots, n$*.*

Based on proximities a background knowledge can be constructed which signify some information about the proximity of terms and predicate symbols.

**Definition 15** *Let* $G$ *be any set of ground terms and* $S$ *any set of predicate symbols. Let* $RG_{FV}$ *and* $RS_{FV}$ *be any proximity over* $G$ *and* $S$ *respectively. The background knowledge is:*

$$Bk = \{RG_{FV_t} \mid t \in G\} \cup \{RS_{FV_p} \mid p \in S\}$$

## 3.2 Computing of uncertainties

So far two steps was made on the way leading to the concept of multivalued knowledge-base: the concept of a multivalued Datalog program and the concept of background knowledge was defined. Now the question is: how can we connect this program with the background knowledge? How can we deduce to the "synonyms"? For example if $(r(a), (0.8, 0.1))$ is an IFS fact and $RS_F(r,s) = (0.6, 0.3), RG_F(a,b) = (0.7, 0.2)$ then what is the uncertainty of $r(b), s(a)$ or $s(b)$?

To solve this problem a new extended uncertainty function will be introduced. According to this function the uncertainty levels of synonyms can be computed from the levels of original fact and from the proximity values of actual predicates and its arguments. It is expectable that in the case of identity the level must be unchanged, but in other cases it is to be less or equal then the original level or then the proximity values. Furthermore we require this function to be monotonically increasing. This function will be ordered to each atom of a program.

Let $p$ be a predicate symbol with $n$ arguments, then $p/n$ is called the functor of the atom characterized by this predicate symbol.

**Definition 16** *A* kb*-extended uncertainty function of* $p/n$ *is:*

$$\vec{\varphi}_p(\vec{\alpha}, \vec{\lambda}, \vec{\lambda_1}, \ldots, \vec{\lambda_n}) : (\vec{0}_{FV}, \vec{1}_{FV}]^{n+2} \to [\vec{0}_{FV}, \vec{1}_{FV}]$$

*where*

$$\vec{\varphi}_p(\vec{\alpha}, \vec{\lambda}, \vec{\lambda_1}, \ldots, \vec{\lambda_n}) \leq \min(\vec{\alpha}, \vec{\lambda}, \vec{\lambda_1}, \ldots, \vec{\lambda_n}),$$
$$\vec{\varphi}_p(\vec{\alpha}, \vec{1}_{FV}, \vec{1}_{FV}, \ldots, \vec{1}_{FV}) = \vec{\alpha}$$

*and* $\vec{\varphi}_p(\vec{\alpha}, \vec{\lambda}, \vec{\lambda_1}, \ldots, \vec{\lambda_n})$ *is monoton increasing in each argument.*

It is worth to mention that any triangular norm is suitable for kb-extended uncertainty function, for example

$$\vec{\varphi}_{p_1}(\vec{\alpha}, \vec{\lambda}, \vec{\lambda_1}, \ldots, \vec{\lambda_n}) = \min(\vec{\alpha}, \vec{\lambda}, \vec{\lambda_1}, \ldots, \vec{\lambda_n}),$$
$$\vec{\varphi}_{p_2}(\vec{\alpha}, \vec{\lambda}, \vec{\lambda_1}, \ldots, \vec{\lambda_n}) = \min(\vec{\alpha}, \vec{\lambda}, \vec{\lambda_1} \cdots \vec{\lambda_n}),$$

where the product is:

$$(\mu_1, \mu_2) \cdot (\lambda_1, \lambda_2) = (\mu_1 \cdot \lambda_1, \mu_2 \cdot \lambda_2),$$

are kb-extended uncertainty functions, but

$$\vec{\varphi}_{p_3}(\vec{\alpha}, \vec{\lambda}, \vec{\lambda_1}, \ldots, \vec{\lambda_n}) = \vec{\alpha} \cdot \vec{\lambda} \cdot \vec{\lambda_1} \cdots \vec{\lambda_n}$$

is a kb-extended uncertainty function only in the interval valued case.

**Example 17** *Let* $(r(a), (0.8, 0.1))$ *be an IFS fact and* $RS_F(r, s) = (0.6, 0.3)$, $RG_F(a, b) = (0.7, 0.2)$ *and* $\vec{\varphi}_r(\vec{\alpha}, \vec{\lambda}, \vec{\lambda_1}) = \min(\vec{\alpha}, \vec{\lambda}, \vec{\lambda_1})$ *then the uncertainty levels of* $r(b), s(a)$ *and* $s(b)$ *are:*

$$(r(b), (\min(0.8, 1, 0.7), \max(0.1, 0, 0.2))) = (r(b), (0.7, 0.2)),$$
$$(s(a), (\min(0.8, 0.6, 1), \max(0.1, 0.3, 0))) = (s(a), (0.6, 0.3)),$$
$$(s(b), (\min(0.8, 0.6, 0.7), \max(0.1, 0.3, 0.2))) = (s(b), (0.6, 0.3)).$$

We have to order kb-extended uncertainty functions to each predicate of the program. The set of these functions will be the function-set of the program.

**Definition 18** *Let* $P$ *be a multivalued Datalog program, and* $F_P$ *be the set of the program's functors. The function-set of* $P$ *is:*

$$\Phi_P = \{\vec{\varphi}_p(\vec{\alpha}, \vec{\lambda}, \vec{\lambda_1}, \ldots, \vec{\lambda_n}) \mid \forall\ p/n \in F_P.\}$$

### 3.3 Connecting algorithm

Let $\mathsf{P}$ be a multivalued Datalog program, Bk be any background knowledge and $\Phi_\mathsf{P}$ be the function-set of $\mathsf{P}$. The deducing mechanism consist of two alternating part: starting from the fact we determine their "synonyms", then applying the suitable rules another facts are derived, then their "synonyms" are derived and again the rules are applied, etc. To define it in a precise manner the concept of modified consecution transformation will be introduced.

The original consequence transformation is defined over the set of all multivalued sets of $\mathsf{P}$'s Herbrand base, that is over $F(B_\mathsf{P})$. To define the modified transformation's domain, let us extend $\mathsf{P}$'s Herbrand universe with all possible ground terms occurring in background knowledge: this way, we obtain the modified Herbrand universe $modH_\mathsf{P}$. Let the modified Herbrand base $modB_\mathsf{P}$ be the set of all possible ground atoms whose predicate symbols occur in $\mathsf{P} \cup \mathsf{Bk}$ and whose arguments are elements of $modH_\mathsf{P}$. This leads to

**Definition 19** *The modified consequence transformation*

$$modNT_\mathsf{P} : FV(modB_\mathsf{P}) \to FV(modB_\mathsf{P})$$

*is defined as*

$$\begin{aligned} modNT_\mathsf{P}(X) = \{ (q(s_1, \ldots, s_n), \vec{\varphi}_p(\vec{\alpha}_p, \vec{\lambda}_q, \vec{\lambda}_{s_1}, \ldots, \vec{\lambda}_{s_n}) \mid \\ (q, \vec{\lambda}_q) \in RS_{FV_p}; \\ (s_i, \vec{\lambda}_{s_i}) \in RG_{t_i},\ 1 \le i \le n \} \cup X, \end{aligned}$$

*where*

$$(p(t_1, \ldots, t_n) \leftarrow A_1, \ldots, A_k; \vec{I}; \vec{\beta}) \in ground(\mathsf{P}),$$
$$(\,|A_i|, \alpha_{A_i}) \in X,\ 1 \le i \le k, \quad (|A_i| \textit{ is the kernel of } A_i)$$

*and* $\vec{\alpha}_p$ *is computed according to the actual extension of (1).*

It is obvious that this transformation is inflationary over $FV(modB_\mathsf{P})$ and it is monotone if $\mathsf{P}$ is positive.
(A transformation $T$ over a lattice $L$ is inflationary if $X \le T(X)\ \forall X \in L$. $T$ is monotone if $T(X) \le T(Y)$ if $X \le Y$.)

According to [15] an inflationary transformation over a complete lattice has a fixed point moreover a monotone transformation has a least fixed point, so

**Proposition 20** *The modified consequence transformation* $modNT_\mathsf{P}$ *has a fixed point. If* $\mathsf{P}$ *is positive, then this is the least fixed point.*

It can be shown that this fixed point is a model of $\mathsf{P}$, but $\mathsf{lfp}(\mathsf{NT_P}) \subseteq \mathsf{lfp}(\mathsf{modNT_P})$, so it is not a minimal model.

As the modifying of original transformation that is the modifying algorithm has no effect on the order of rules, therefore it does not change the stratification. Therefore we can state

**Proposition 21** *In the case of stratified program* $\mathsf{P}$, $\mathsf{modNT_P}$ *has least fixed point as well.*

Now we have all components together to define the concept of a multivalued knowledge-base. But before doing it, it is worth mentioning that the above modified consequence transformation is not the unique way to connect the background knowledge with the deduction mechanism, there could be other possibilities as well.

**Definition 22** *A multivalued knowledge-base (*$\mathsf{mKB}$*) is a quadruple*

$$\mathsf{mKB} = (\mathsf{Bk}, \mathsf{P}, \Phi_\mathsf{P}, \mathsf{cA}),$$

*where* $\mathsf{Bk}$ *is a background knowledge,* $\mathsf{P}$ *is a multivalued Datalog program,* $\Phi_\mathsf{P}$ *is a function-set of* $\mathsf{P}$ *and* $\mathsf{cA}$ *is any connecting algorithm.*

The result of the connected and evaluated program is called the consequence of the knowledge-base, denoted by

$$\mathsf{C}(\mathsf{Bk}, \mathsf{P}, \Phi_\mathsf{P}, \mathsf{cA}).$$

So in our case $\mathsf{C}(\mathsf{Bk}, \mathsf{P}, \Phi_\mathsf{P}, \mathsf{cA}) = \mathsf{lfp}(\mathsf{modNT_P})$.

**Example 23** *Let the IVS valued* $\mathsf{mDATALOG}$ *program and the background knowledge be as follows*

$$\mathsf{lo}(x, y) \leftarrow \mathsf{gc}(y), \mathsf{mu}(x); (0.7, 0.9); \vec{\mathsf{I}}_{\mathsf{VG}}.$$
$$(\mathsf{fv}(\mathsf{V}), (0.85, 0.9).$$
$$(\mathsf{mf}(\mathsf{M}), (0.7, 0.8).$$

| | *B* | *V* | *M* |
|---|---|---|---|
| *B* | (1, 1) | (0.8, 0.9) | |
| *V* | (0.8, 0.9) | (1, 1) | |
| *M* | | | (1, 1) |

| | lo | li | gc | fv | mu | mf |
|---|---|---|---|---|---|---|
| lo | $(1,1)$ | $(0.7,0.9)$ | | | | |
| li | $(0.7,0.9)$ | $(1,1)$ | | | | |
| gc | | | $(1,1)$ | $(0.8,0.9)$ | | |
| fv | | | $(0.8,0.9)$ | $(1,1)$ | | |
| mu | | | | | $(1,1)$ | $(0.6,0.7)$ |
| mf | | | | | $(0.6,0.7)$ | $(1,1)$ |

*According to the connecting algorithm, it is enough to consider only the extended uncertainty functions of head-predicates. Let these functions be as follows:*

$$\begin{aligned}
\vec{\varphi}_{lo}(\vec{\alpha},\vec{\lambda},\vec{\lambda_1},\vec{\lambda_2}) &:= \min(\vec{\alpha},\vec{\lambda},\vec{\lambda_1}\cdot\vec{\lambda_2}),\\
\vec{\varphi}_{fv}(\vec{\alpha},\vec{\lambda},\vec{\lambda_1}) &:= \min(\vec{\alpha},\vec{\lambda},\vec{\lambda_1}),\\
\vec{\varphi}_{mf}(\vec{\alpha},\vec{\lambda},\vec{\lambda_1}) &:= \vec{\alpha}\cdot\vec{\lambda}\cdot\vec{\lambda_1}.
\end{aligned}$$

*The modified consequence transformation takes shape in the following steps:*

$$X_0 = \{(fv(V),(0.85,0.9)),(mf(M),(0.7,0.8))\}$$

$$\Downarrow \quad \textit{(according to the proximity)}$$

$$\begin{aligned}
X_1 = modNT_P(X_0) &= X_0\cup\\
\{(gc(V), &\vec{\varphi}_{fv}((0.85,0.9),(0.8,0.9),(1,1)) =\\
&(\min(0.85,0.8,1),\min(0.9,0.9,1)) = (0.8,0.9)),\\
&(fv(B),\vec{\varphi}_{fv}((0.85,0.9),(1,1),(0.8,0.9)) = (0.8,0.9)),\\
&(gc(B),\vec{\varphi}_{fv}((0.85,0.9),(0.8,0.9),(0.8,0.9)) = (0.8,0.9)),\\
&(mu(M),\vec{\varphi}_{mf}((0.7,0.8),(0.6,0.7),(1,1)) =\\
&(0.7\cdot 0.6\cdot 1, 0.8\cdot 0.7\cdot 1) = (0.42,0.56))\}
\end{aligned}$$

$$\Downarrow \quad \textit{(applying the rules)}$$

$$\begin{aligned}
lo(M,V) &\leftarrow gc(V),mu(M);(0.7,0.9);\vec{I}_{VG}.\\
lo(M,B) &\leftarrow gc(B),mu(M);(0.7,0.9);\vec{I}_{VG}.\\
&\textit{here}: f_V(\vec{I}_{VG},\vec{\alpha},\vec{\beta}) = \min(\vec{\alpha}_{body},\vec{\beta}),\ \textit{so}
\end{aligned}$$

$$\begin{aligned}
X_2 = modNT_P(X_1) &= X_1\cup\\
\{(lo(M,V),(0.42,0.56)),&(lo(M,V),(0.42,0.56))\}
\end{aligned}$$

$$\Downarrow \quad \textit{(according to the proximity)}$$

$$\begin{gathered}
X_3 = \mathsf{modNT_P}(X_2) = X_2 \cup \\
\{(li(M,V),(min(0.42,0.7,1\cdot 1),min(0.56,0.9,1\cdot 1))), \\
(li(M,B),(min(0.42,0.7,1\cdot 1),min(0.56,0.9,1\cdot 1)))\} \cup \\
\{(li(M,V),(min(0.42,0.7,0.64),min(0.56,0.9,0.81))), \\
(li(M,B),(min(0.42,0.7,0.64),min(0.56,0.9,0.81)))\}
\end{gathered}$$

$X_3$ *is a fixed point, so the consequence of the knowledge-base is:*

$$\begin{aligned}
&C(Bk,P,\Phi_P,cA) = \\
&\{(fv(V),(0.85,0.9)),(mf(M),(0.7,0.8)), \\
&(gc(V),(0.8,0.9)),(fv(B),(0.8,0.9)), \\
&(gc(B),(0.8,0.9)),(mu(M),(0.42,0.56)), \\
&(lo(M,V),(0.42,0.56)),(lo(M,V),(0.42,0.56)) \\
&(li(M,V),(0.42,0.56)),(li(M,B),(0.42,0.56))\}
\end{aligned}$$

To illustrate our discussion with some realistic content, in the above example the knowledge-base could have the following interpretation. Let us suppose that music listeners "generally" (level between 0.7, 0.9) are fond of the greatest composers. Assume furthermore that Mary is a "rather devoted" (level between 0.7, 0.8) fan of classical music (mf), and Vivaldi is "generally accepted" (level between 0.85, 0.9) as a "great composer". It is also widely accepted that the music of Vivaldi and Bach are fairly "similar", being related in overall structure and style. On the basis of the above information, how strongly state that Mary likes Bach? To continue with this idea, next we can assume that an internet agent wants to suggest a good CD for Mary, based on her interests revealed through her actions at an internet site. A multivalued knowledge-base could help the agent to get a good answer. As some of the readers may well know, similar mechanisms – but possibly based on entirely different modelling paradigms – are in place in prominent websites such as Amazon and others.

## 4 Evaluation strategies

As the above example shows (especially in the case of enlarging the program with other facts and rules), the fixed point-query may involve many superfluous calculations, because sometimes we want to give an answer for a concrete question, and we are not interested in the whole sequence.

If a goal (query) is specified together with the multivalued knowledge-base, then it is enough to consider only the rules and facts being necessary to reach the goal. In this section we deal with a possible evaluation of knowledge-base, which is a combination of top-down and bottom-up evaluation.

A goal is a pair $(q(t_1, t_2, \ldots t_n), \vec{\alpha})$, where $(q(t_1, t_2, \ldots t_n)$ is an atom, $\vec{\alpha}$ is the fuzzy, the intuitionistic, the interval-valued or the bipolar level of the atom. $q$ may contain variables, and its levels may be known or unknown values.

According to the top down evaluation a goal is evaluated through subqueries. This means that there are selected all possible rules, whose head can be unified with the given goal, and the atoms of the body are considered as new sub-goals. This procedure continues until the facts are obtained.

The top-down evaluation of a multivalued Datalog program does not terminate by obtaining the facts, because we need to determine the uncertainty levels of the goal. The algorithm given in [5, 8] calculates this level in a bottom-up manner: starting from the leaves of the evaluating graph, going backward to the root, and applying the uncertainty-level functions along the suitable path of this graph, finally we get the uncertainty level of the root.

In the case of knowledge-base, we rely on the bottom-up evaluation, but the selection of required starting facts takes place in a top-down fashion. Since only the required starting facts are sought, in the top-down part of the evaluation there is no need for the uncertainty levels. Hence, we search only among the ordinary facts and rules. To do this, we need the concept of substitution and unification which are given for example in [8, 15, 21], etc.

But now sometime we also need other kinds of substitutions: to substitute some predicate $p$ or term $t$ for their proximity sets $R_p$ and $R_t$, and to substitute some proximity sets for their members.

Next, for the sake of simpler terminology, we mean by goal, rules and facts these concepts without uncertainty levels. An AND/OR tree arises during the evaluation, this is the searching tree. Its root is the goal; its leaves are either YES or NO. The parent nodes of YES are the required starting facts. This tree is build up by alternating proximity-based and rule-based unification.

The proximity-based unification unifies the predicate symbols of sub-goals by the members of its proximity set, except the first and last unification. The first proximity-based unification unifies the ground terms of the goal with their proximity sets, and the last one unifies the proximity sets among the parameters of resulting facts with their members.

The rule-based unification unifies the sub-goals with the head of suitable rules, and continues the evaluating by the bodies of these rules. During this unification the proximity sets of terms are considered as ordinary constants, and a constant can be unify with its proximity set. The searching graph according to its depth is build up in the following way:

If the goal is on depth 0, then every successor of any node on depth $3k+2(k = 0, 1, \ldots)$ is in AND connection, the others are in OR connection. In detail:

The successors of a goal $q(t_1, t_2, \ldots t_n)$ will be all possible $q'(t'_1, t'_2, \ldots, t'_n)$, where $q' \in R_q$; $t'_i = t_i$ if $t_i$ is some variable and $t'_i = R_{t_i}$ if $t_i$ is a ground term.

If the atom $p(t_1, t_2, \ldots t_n)$ is in depth $3k(k = 1, 2, \ldots)$, then the successor nodes be all possible $p'(t_1, t_2, \ldots t_n)$, where $p' \in R_p$.

If the atom L is in depth $3k + 1(k = 1, 2, \ldots)$, then the successor nodes will be the bodies of suitable unified rules, or the unified facts, if L is unifiable with any fact of the program, or NO, if there is not any unifiable rule or fact. That is, if the head of rule $M \leftarrow M_1, \ldots, M_n, (n > 0)$ is unifiable with $L$, then the successor of $L$ be $M_1\theta, \ldots, M_n\theta$, where $\theta$ is the most general unification of $L$ and $M$.

If $n = 0$, that is in the program there is any fact with the predicate symbol of $L$, then the successors be the unified facts. If $L = p(t_1, t_2, \ldots, t_n)$ and in the program there is any fact with predicate symbol $p$, then the successor nodes be all possible $p(t'_1, t'_2, \ldots, t'_n)$, where $t'_i \in R_{t_i}$ if $t_i = R_{t_i}$ or $t'_i = t_i\theta$, if $t_i$ is a variable, and $\theta$ is a suitable unification.

According the previous paragraph, there are three kinds of nodes in depth $3k + 2(k = 1, 2, \ldots)$: a unified body of a rule; a unified fact with ordinary ground term arguments; or the symbol NO.

In the first case the successors are the members of the body. They are in AND connection, which is not important in our context, but maybe important for possible future development. If the body has only one literal, then the length of evaluating path would be reduced to one, but it would "damage" the view of homogeneous treatment. In the second case the successors are the symbol YES or NO, depending on whether the unified fact is among the ground atoms of the program. The NO-node has not successor.

From the construction of searching graph, we conclude

**Proposition 24** *Let $X_0$ be the set of ground facts being in parent-nodes of symbols YES. Starting from $X_0$, the fixed point of $mNT_P$ contains the answer for the query.*

From the viewpoint of the query, this fixed point may contain more superfluous ground atom, but generally it is smaller then the consequence of knowledge-base. More reduction of the number of superfluous resulting facts is the work of a possible further development.

**Example 25** *Let us consider the knowledge-base of example 23. (Now it is enough to consider only the program and the background knowledge.)*
*Let the goal be:*

*a/ li(M,B).*
*b/ li(M,x), where x is a variable.*

*Then the searching graphs are:*

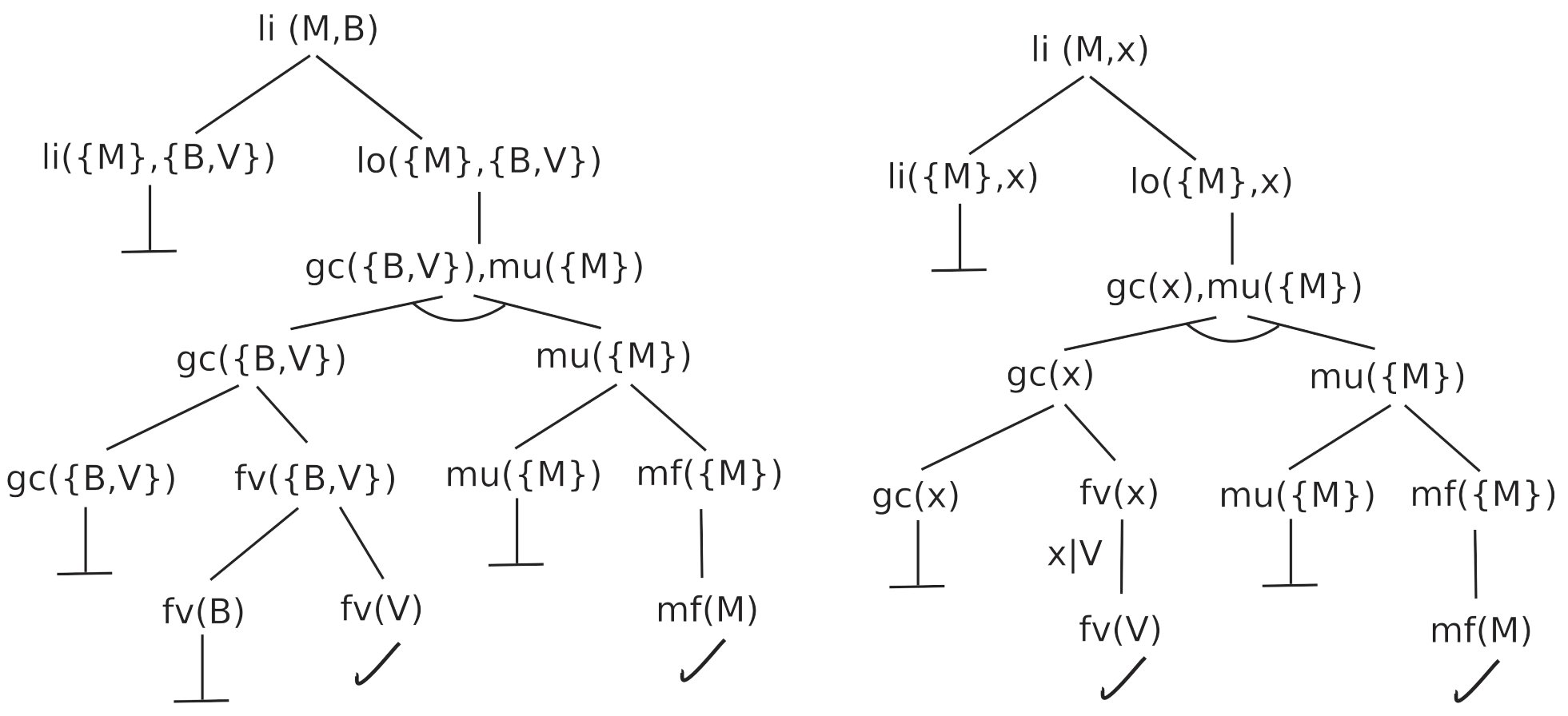

According to the above construction, the searching algorithm is the following alternation of proximity-based and rule-based unification.

## Algorithm

procedure evaluation(g($\underline{\mathsf{t}}$)) /* g($\underline{\mathsf{t}}$) *is the goal* */
  Heads := {the heads of the program's rules}
  Facts := {the facts of the program}
  Resulting_Facts := $\emptyset$ /* *the set of resulting starting facts* */
  for all $\mathsf{t} \in \underline{\mathsf{t}}$ do
    if is_variable($\mathsf{t}$) then $\mathsf{s} := \mathsf{t}$
      else $\mathsf{s} := \mathsf{St}$ /* $\mathsf{St}$ *is the proximity set of* $\mathsf{t}$ */
    end_if
  end_for

  Nodes := {$\mathsf{g}(\underline{\mathsf{s}})$} /* Nodes *is the set of evaluable nodes,* $\underline{\mathsf{s}}$ *is the vector of elements* $\mathsf{s}$ *in the original order* */

```
  New_nodes := ∅                /* the successor nodes of Nodes */
  while not_empty(Nodes) do
    p(t) := element(Nodes)
    Spnodes := ∅                /* the successor nodes of p(t) */
    proximity_evaluation(p(t),Spnodes)
    New_nodes := New_nodes ∪ Spnodes
    Nodes := Nodes − {p(t)}
  end_while
  Nodes := New_nodes
  New_nodes := ∅
  while not_empty(Nodes) do
    p(t) := element(Nodes)
    Spnodes := ∅                /* the successor nodes of p(t) */
    rule_evaluation(p(t),Spnodes)
    New_nodes := New_nodes ∪ Spnodes
    Nodes := Nodes − {p(t)}
  end_while
  return Resulting_Facts
end_procedure

procedure proximity_evaluation(p(t),Spnodes)
  for all q ∈ Sp do             /* Sp is the proximity set of p */
    Spnodes := Spnodes ∪ {q(t)}
  end_for
end_procedure

procedure rule_evaluation(p(t),Spnodes)
  for all p(v) ∈ Heads do
    if is_unifiable(p(t),p(v)) then
    Spnodes := Spnodes ∪
                      {unified predicates of the body belonging to p(vθ)}
                           /* θ is the suitable unifier */
    end_if
  end_for
  for all p(v) ∈ Facts do
    if is_unifiable(p(t),p(v)) then
    for all St ∈ vθ do          /* θ is the suitable unifier */
      if is_variable(St) then
      t := Stτ                  /* τ is the suitable unifier */
```

```
        else if is_proximity_set(St) then
        t := element(St)
        end_if
      end_for
      end_if
      for all possible t̲ do
                                /* t̲ is the vector of elements t in the right order */
        if p(t̲) ∈ Facts then
          Resulting_Facts := Resulting_Facts ∪ {p(t̲)}
        end_if
      end_for
    end_for
  end_procedure
```

This algorithm can be applied for stratified multivalued Datalog too by determining the successor of a rule-body without negation.

## 5 Conclusions

In this paper we have presented a model of handling uncertain information by defining the multivalued knowledge-base as a quadruple of background knowledge, a deduction mechanism, a decoding set and a modifying algorithm which connects the background knowledge to the deduction mechanism. We also have presented a possible evaluation strategy. To improve upon this strategy and/or the modifying algorithm and/or the structure of background knowledge will be the subject of further investigations. An efficient multivalued knowledge base could be the basis of decisions based on uncertain information, or would be a possible method for handling argumentation or negotiation of agents.

## References

[1] Á. Achs, A. Kiss, Fixpoint query in fuzzy Datalog, *Ann. Univ. Sci., Budapest., Sect. Comput.* **15** (1995) 223–231. ⇒51, 54, 55

[2] Á. Achs, A. Kiss, Fuzzy extension of Datalog, *Acta Cybernet.* (Szeged), **12,** 2 (1995) 153–166. ⇒51, 55

[3] Á. Achs, Creation and evaluation of fuzzy knowledge-base, *J. UCS*, **12,** 9 (2006) 1087–1103. ⇒52

[4] Á. Achs, Twofold extensions of fuzzy datalog, *7th International Workshop on Fuzzy Logic and Applications, WILF 2007*, Camogli, Italy, 2007, pp. 298–306. ⇒ 52

[5] Á. Achs, Computed answer from uncertain knowledge: A model for handling uncertain information, *Comput. Inform.*, **26,** 1 (2007) 298–306. ⇒ 73

[6] Á. Achs, DATALOG-based uncertainty-handling, *The Twelfth IASTED International Conference on Artificial Intelligence and Soft Computing*, Mallorca, Spain, 2008, pp. 44–49. ⇒ 52

[7] Á. Achs, Multivalued Knowledge-Base based on multivalued Datalog, *Proc. World Academy of Science, Engineering and Technology*, **54** (2009) 160–165. ⇒ 52

[8] Á. Achs, Fuzziness and intuitionism in database theory, in: *Intuitionistic Fuzzy Sets: Recent Advances*, to appear ⇒ 66, 73

[9] K. Atanassov, Intuitionistic fuzzy sets, *VII ITKR's Session, Sofia (deposed in Central Science-Technical Library of Bulgarian Academy of Science, 1697/84)*, 1983. ⇒ 56

[10] K. Atanassov, G. Gargov, Interval-valued intuitionistic fuzzy sets, *Fuzzy Sets and Systems*, **31,** 3 (1989) 343–349. ⇒ 56

[11] K. Atanassov, *Intuitionistic fuzzy sets*, Springer-Verlag, Heidelberg, 1999. ⇒ 56

[12] K. Atanassov, Intuitionistic fuzzy implications and Modus Ponens, *Notes on Intuitionistic Fuzzy Sets*, **11** (2005) 1–5. ⇒ 58

[13] K. Atanassov, On some intuitionistic fuzzy implications, *C. R. Acad. Bulgare Sci.*, **59** (2006) 19–24. ⇒ 58

[14] J. T. Cacioppo, W. L. Gardner, G. G. Berntson, Beyond bipolar conceptualization and measures: the case of attitudes and evaluative spaces, *Personality and Social Psychol. Rev.*, **1** (1997) 3–25. ⇒ 62

[15] S. Ceri, G. Gottlob, L. Tanca, *Logic Programming and Databases*, Springer–Verlag, Berlin, 1990. ⇒ 53, 57, 69, 73

[16] C. Cornelis, G. Deschrijver, E. E. Kerre, Implication in intuitionistic fuzzy and interval-valued fuzzy set theory: construction, classification, application, *Internat. J. Approx. Reason.*, **35** (2004) 55–95. ⇒57, 58

[17] D. Dubois, H. Prade, Fuzzy sets in approximate reasoning, Part 1: Inference with possibility distributions, *Fuzzy Sets and Systems*, **40** (1991) 143–202. ⇒55

[18] D. Dubois, P. Hajek, H. Prade, Knowledge-Driven versus Data-Driven Logics, *J. Log. Lang. Inf.*, **9** (2000) 65–89. ⇒62

[19] D. Dubois, S. Gottwald, P. Hajek, J. Kacprzyk, H. Prade, Terminological difficulties in fuzzy set theory – The case of "Intuitionistic Fuzzy Sets", *Fuzzy Sets and Systems*, **15** (2005) 485–491. ⇒62

[20] J. W. Lloyd, *Foundations of Logic Programming*, Springer–Verlag, Berlin, 1990. ⇒53

[21] J. D. Ullman, *Principles of Database and Knowledge-base Systems*, Computer Science Press, Rockville, 1988. ⇒53, 73